\documentclass[letterpaper, 10 pt, journal, twoside]{ieeetran} 

\IEEEoverridecommandlockouts                              

\usepackage{amsfonts}	
\usepackage{amsmath}	
\usepackage{amssymb}    
\usepackage{xfrac}    
\usepackage{siunitx}
\usepackage{mathtools}
\usepackage{pifont}   
\usepackage{dsfont}   

\usepackage{cuted} 

\usepackage{booktabs}
\usepackage{makecell}  
\usepackage[flushleft]{threeparttable}  
\usepackage{multirow}
\usepackage{colortbl}

\usepackage{xspace}    
\usepackage[table]{xcolor}
\usepackage{xcolor}    
\usepackage{textcomp}  
\usepackage{gensymb}   
\usepackage{graphicx} 

\makeatletter
\let\NAT@parse\undefined
\makeatother

\usepackage[pdfencoding=auto, hidelinks]{hyperref} 
\usepackage[hang,flushmargin]{footmisc}

\usepackage[capitalize]{cleveref}
\crefname{section}{Sec.}{Secs.}
\Crefname{section}{Section}{Sections}
\Crefname{table}{Table}{Tables}
\crefname{table}{Tab.}{Tabs.}

\newcommand{\net}{\mbox{PASTEL}\xspace}

\newcommand{\greyrule}{\arrayrulecolor{black!30}\midrule\arrayrulecolor{black}}

\definecolor{Gray}{gray}{0.9}%

\DeclareMathOperator*{\argmax}{arg\,max}

\DeclareMathOperator*{\pline}{line}

\begin{document}

\title{A Good Foundation is Worth Many Labels: Label-Efficient Panoptic Segmentation}

\author{
Niclas Vödisch$^{1*}$,
Kürsat Petek$^{1*}$,
Markus Käppeler$^{1*}$,
Abhinav Valada$^{1}$, and
Wolfram Burgard$^{2}$
\thanks{$^{*}$ Equal contribution.}%
\thanks{© 2024 IEEE. Personal use of this material is permitted. Permission from IEEE must be obtained for all other uses, in any current or future media, including reprinting/republishing this material for advertising or promotional purposes, creating new collective works, for resale or redistribution to servers or lists, or reuse of any copyrighted component of this work in other works.}
\thanks{This work was funded by the German Research Foundation Emmy Noether Program grant No 468878300.}%
\thanks{$^{1}$ Niclas Vödisch, Kürsat Petek, Markus Käppeler, and Abhinav Valada are with the Department of Computer Science, University of Freiburg, Germany.}%
\thanks{$^{2}$ Wolfram Burgard is with the Department of Engineering, University of Technology Nuremberg, Germany.}%
\thanks{Digital Object Identifier (DOI): \href{https://www.doi.org/10.1109/LRA.2024.3505779}{10.1109/LRA.2024.3505779}}
}

\markboth{T\MakeLowercase{his paper appeared in:} IEEE ROBOTICS AND AUTOMATION LETTERS, VOL. 10, ISSUE 1, JANUARY 2025}%
{Vödisch \MakeLowercase{\textit{et al.}}: A Good Foundation is Worth Many Labels: Label-Efficient Panoptic Segmentation}

\maketitle


\begin{abstract}
    A key challenge for the widespread application of learning-based models for robotic perception is to significantly reduce the required amount of annotated training data while achieving accurate predictions. This is essential not only to decrease operating costs but also to speed up deployment time. In this work, we address this challenge for \textbf{PA}noptic \textbf{S}egmen\textbf{T}ation with f\textbf{E}w \textbf{L}abels (\net) by exploiting the groundwork paved by visual foundation models. We leverage descriptive image features from such a model to train two lightweight network heads for semantic segmentation and object boundary detection, using very few annotated training samples. We then merge their predictions via a novel fusion module that yields panoptic maps based on normalized cut. To further enhance the performance, we utilize self-training on unlabeled images selected by a feature-driven similarity scheme. We underline the relevance of our approach by employing \net to important robot perception use cases from autonomous driving and agricultural robotics. In extensive experiments, we demonstrate that \net significantly outperforms previous methods for label-efficient segmentation even when using fewer annotations. The code of our work is publicly available at \mbox{\url{http://pastel.cs.uni-freiburg.de}}.

\end{abstract}




\section{Introduction}
\label{sec:introduction}

\IEEEPARstart{H}{olistic} scene understanding is a core requirement for mobile robots to interact autonomously with their environment. Commonly, this is addressed by visual panoptic segmentation that assigns a semantic class to each pixel while separating instances of the same class.
Although recent methods~\cite{cheng2020panoptic, cheng2022mask2former, mohan2022perceiving} have shown great progress in terms of segmentation performance, they often rely on a vast amount of densely annotated training data and tend to generalize poorly to new domains. Since creating panoptic labels is a highly laborious task~\cite{cordts2016cityscapes}, collecting large-scale training data for every new area of operation would drastically increase the cost of robot deployment. This particularly hinders the widespread application in continuously changing environments, e.g., agricultural robotics.
To reduce training costs, some recent segmentation techniques employ various kinds of limited supervision. For instance, by learning from sparse annotations~\cite{li2023point2mask, li2023panopticfcn}, in semi-~\cite{hoyer2021three, yang2022stpp, wang2022u2pl} or unsupervised manners~\cite{hyun2021picie, wang2023cut}, and more recently by leveraging foundation models~\cite{hamilton2022stego, kaeppeler2023spino}. Since these models can be adapted to various downstream tasks~\cite{bommasani2021opportunities, oquab2023dinov2}, we argue that they offer a powerful pretraining strategy for addressing robotic perception tasks in a label-efficient manner.\looseness=-1

\begin{figure}[t]
    \centering
    \includegraphics[width=.9\linewidth]{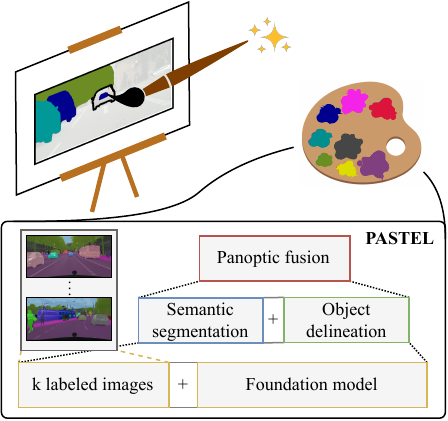}
    \vspace{-.2cm}
    \caption{We propose PASTEL for label-efficient panoptic segmentation. Our method combines a DINOv2~\cite{oquab2023dinov2} backbone, creating descriptive image features, with labels from only $k$ images, e.g., $k=10$ on Citycapes~\cite{cordts2016cityscapes}. A novel fusion module then merges semantic predictions with estimated object boundaries to yield the panoptic output.}
    \vspace{-.5cm}
    \label{fig:teaser}
\end{figure}

In this work, we employ this paradigm shift to panoptic segmentation to substantially reduce the number of annotated images required for training.
In particular, we propose a novel approach for \textit{PAnoptic SegmenTation with fEw Labels}~(\net) and illustrate the key idea in \cref{fig:teaser}.
First, the vision foundation model DINOv2~\cite{oquab2023dinov2} and a small set of $k$ densely annotated images form the basis of \net. Second, the descriptive image features of DINOv2~\cite{oquab2023dinov2} allow for highly label-efficient training of two lightweight heads for semantic segmentation and object boundary estimation. Third, at inference time, a novel panoptic fusion module then merges the task-specific predictions and further refines their quality.
Finally, \net bootstraps selectively sampled unlabeled images for an additional performance boost via self-training.
In extensive experiments, we demonstrate that \net creates high-quality panoptic predictions from as few as 10 labeled images on Cityscapes~\cite{cordts2016cityscapes}, Pascal VOC~\cite{everingham2010pascal}, and PhenoBench~\cite{weyler2024phenobench}. Notably, \net can hence be trained with labels produced by a single annotator in 1\sfrac{1}{2} days~\cite{cordts2016cityscapes} while outperforming previous label-efficient methods that require five to ten times as much data. We further show that the predictions of \net can be used as pseudo-labels to train densely supervised models, i.e., rendering them label-efficient. To encourage future research, we release our code at \mbox{\url{http://pastel.cs.uni-freiburg.de}}.

\section{Related Work}
\label{sec:related-work}

We provide an overview of visual foundation models and previous methods for label-efficient image segmentation.


{\parskip=3pt
\noindent\textit{Visual Foundation Models.}
The term ``foundation model'' defines models that are trained on large amounts of data for adaptation to a variety of downstream tasks~\cite{bommasani2021opportunities}.
First applied in natural language processing, e.g., GPT-3~\cite{brown2020gpt3}, similar approaches have since also been proposed for computer vision (CV).
For instance, CLIP~\cite{radford2021clip} allows for zero-shot image classification that can be leveraged in open-vocabulary methods~\cite{lin2023clipes}. Florence~\cite{yuan2021florence} represents a general-purpose CV foundation model by extending the textual-visual shared representation to the space and time domains.
Similarly, Painter~\cite{wang2023painter} addresses common CV tasks such as image segmentation or depth estimation without task-specific heads.
The recent SAM~\cite{kirillov2023sam} enables zero-shot semantic and instance segmentation while lacking the ability to assign classes to the segmented areas.
Finally, DINO~\cite{caron2021dino} and DINOv2~\cite{oquab2023dinov2} represent a new paradigm of visual foundation models relying on a completely unsupervised training scheme with neither cross-modal nor iterative human annotations. Nonetheless, these models have been shown to learn semantically descriptive features for downstream tasks~\cite{oquab2023dinov2, kaeppeler2023spino}.
In this work, we exploit such image representations as a strong prior to enable label-efficient panoptic segmentation.
}


{\parskip=3pt
\noindent\textit{Label-Efficient Image Segmentation.}
Classical deep image segmentation methods require a large amount of annotated training data~\cite{cheng2020panoptic, cheng2022mask2former, mohan2022perceiving}. Therefore, many recent works employ different strategies of weak supervision~\cite{shen2023asurvey} to reduce the labeling cost.
For instance, unsupervised semantic segmentation is commonly addressed using contrastive learning techniques~\cite{hyun2021picie, gansbeke2021maskcontrast} to find similar clusters in the feature space. Recent methods have leveraged descriptive image representations from large-scale task-agnostic pretraining~\cite{caron2021dino} for both semantic~\cite{hamilton2022stego} and instance segmentation~\cite{wang2023cut}.
With respect to the more challenging task of panoptic segmentation, CoDEPS~\cite{voedisch23codeps} distills knowledge from a labeled source domain to a new unlabeled target domain.
Sparse annotations offer an intermediate approach with limited pixel-based supervision generated by an inexpensive labeling scheme, e.g., point annotations~\cite{li2023point2mask, li2023panopticfcn}.
Semi-supervised training bootstraps a few densely annotated examples with a large set of unlabeled images and is commonly based on auxiliary tasks~\cite{hoyer2021three}, self-training~\cite{yang2022stpp}, or uncertainty estimation~\cite{wang2022u2pl}. In this work, we add to the promising line of research that exploits visual foundation models for label-efficient dense supervision. The concurrent SPINO~\cite{kaeppeler2023spino} approach combines a DINOv2~\cite{oquab2023dinov2} backbone with separate heads for semantic segmentation and object boundary estimation. Inspired by these recent insights, we follow a similar design scheme but exploit further synergies between semantic classes and leverage unlabeled images via self-training.
Importantly, unlike most prior label-efficient techniques, in this work we aim to enable panoptic segmentation from only as few labeled images as a single annotator can produce within a reasonable time frame, thus facilitating deployment in custom domains.
}

\section{Technical Approach}
\label{sec:technical-approach}

In this section, we present our \net approach for label-efficient panoptic segmentation including its network architecture, the training scheme, the novel panoptic fusion module, and the feature-driven iterative self-training.

\begin{figure*}[t]
    \centering
    \includegraphics[width=\linewidth]{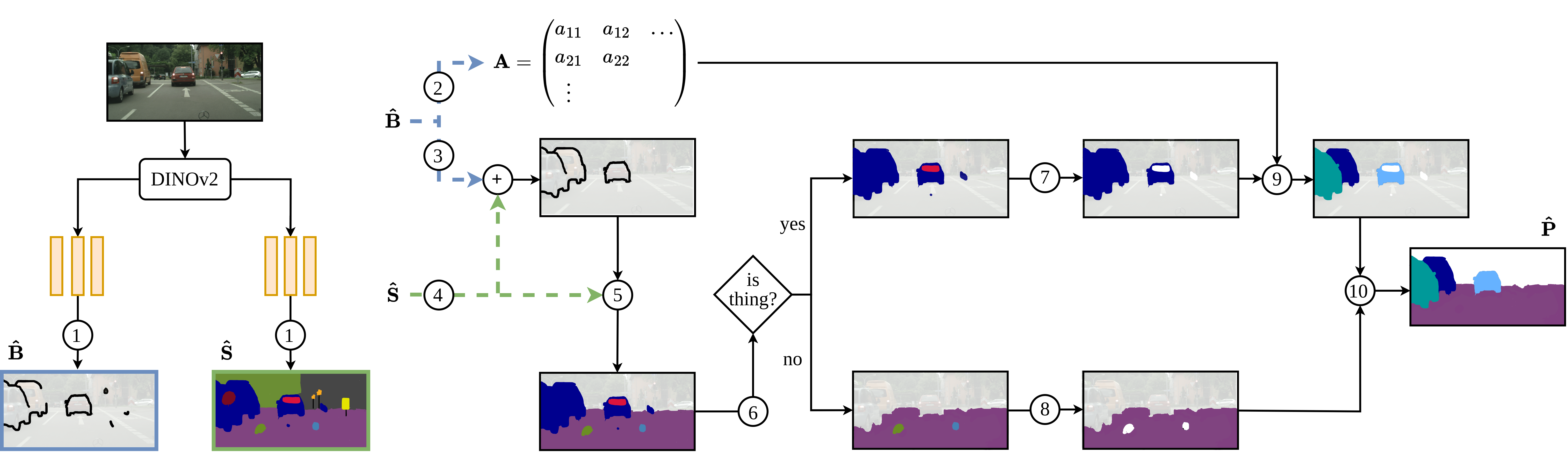}
    \vspace*{-.7cm}
    \caption{Test-time overview of \net illustrating the panoptic fusion scheme. For simplicity, we focus on \textit{car} and \textit{road} classes after step (1). The overall module is comprised of the following steps:
    (1)~Overlapping multi-scale predictions;
    (2)~Conversion of soft boundary map to an affinity matrix;
    (3)~Boundary denoising;
    (4)~Extraction of ``stuff'' to ``thing'' boundaries;
    (5)~Class majority voting within enclosed areas;
    (6)~Connected component analysis (CCA);
    (7)~Filters on ``thing'' classes;
    (8)~Filters on ``stuff'' classes;
    (9)~Recursive two-way normalized cut (NCut) to separate connected instances;
    (10)~Nearest neighbors-based hole filling of pixels with the \textit{ignore} class.
    }
    \label{fig:overview}
    \vspace*{-.5cm}
\end{figure*}


\subsection{Model Architecture and Training}

The key insight of our \net is to exploit the semantically rich image features from a foundation model to enable label-efficient segmentation and instance delineation.

{\parskip=3pt
\noindent\textit{Network Design.}
As illustrated in Fig.~2, we design our network according to the multi-task paradigm with a shared backbone.
Inspired by the approaches Point2Mask~\cite{li2023point2mask} and SPINO~\cite{kaeppeler2023spino}, we separately perform pixel-based semantic segmentation and object boundary detection while using a shared backbone. In detail, we employ the pretrained \mbox{ViT-B/14} variant of DINOv2~\cite{oquab2023dinov2} as the frozen backbone. In the $n$-class segmentation head, we first upsample the patch-wise features of DINOv2 to the input image size, i.e., $14\times$-upsampling.
We then feed the output to four $1\times$1 convolution layers of feature sizes $300$, $300$, $200$, and $n$.
In the object boundary head, we operate on a smaller feature map using a $4\times$-upsampling layer, again followed by four $1\times$1 convolution layers of output sizes $600$, $600$, $400$, and $1$. We frame the boundary detection task as binary classification with labels $0$ and $1$ denoting boundary and background pixels, respectively. During test-time, the output of both heads is merged by our novel panoptic fusion module as detailed in \cref{ssec:fusion-module}.
}

{\parskip=3pt
\noindent\textit{Network Training.}
Due to the descriptive image features of the DINOv2~\cite{oquab2023dinov2} backbone, we can train both heads with a minimum number $k$ of annotated images. In practice, $k$ can be as small as ten samples as shown in \cref{sec:evaluation}.
To train the semantic segmentation head, we employ the bootstrapped cross-entropy loss~\cite{pohlen2017full} to compensate for an imbalanced class distribution:\looseness=-1
\begin{equation}
    \mathcal{L}_\mathit{sem} = \frac{-1}{K} \sum_{i=1}^N \mathds{1} \left[ p_{i, y_i} < t_K \right] \cdot \log (p_{i, y_i}) \, ,
    \label{eqn:sem-loss}
\end{equation}
where $N$ denotes the number of pixels. Furthermore, $p_{i, y_i}$ refers to the posterior probability of pixel~$i$ for the true class \mbox{$y_i \in \{ 1, \dots, n \}$}. Note that $n$ corresponds to the number of semantic classes. The indicator function~$\mathds{1}(\cdot)$ returns $1$ if $p_{i, y_i}$ is smaller than the threshold $t_K$ and $0$ otherwise. To bootstrap pixels with yet uncertain predictions, i.e., a high loss, we set $t_K = 0.2$.
Since we formulate the boundary detection as a $2$-class classification task, we supervise this head with the binary cross entropy loss:
\begin{equation}
    \mathcal{L}_\mathit{bnd} = \frac{-1}{N} \sum_{i=1}^N y_i \cdot \log (p_{i}) + (1 - y_i) \cdot \log (1 - p_{i}) \, ,
\end{equation}
where $N$ is the number of pixels, $y_i \in \{0, 1\}$ denotes the binary boundary label, and $p_{i}$ refers to the pixel probability of being a boundary. During training, we set the true $y_i$ to~$0$ if the instance identifier of a ``thing'' pixel differs from the identifier of any of its eight neighbors. Otherwise, we assign~$1$. If the pixel $i$ belongs to a ``stuff'' class, we set $y_i = 1$.
\looseness=-1

In order to increase the variety of the small training set of only $k$ samples, we employ extensive data augmentation. In particular, we perform randomized horizontal flipping and cropping with consecutive resizing to the input image size. We further augment various visual properties including brightness, contrast, saturation, and hue value.
}


\subsection{Panoptic Fusion Module}
\label{ssec:fusion-module}

Our proposed panoptic fusion module comprises three key steps: generating multi-scale predictions, a variety of heuristic-driven refinements, and the final instance delineation. We illustrate the overall methodology in \cref{fig:overview}.


\begin{figure}[t]
    \centering
    \includegraphics[width=\linewidth]{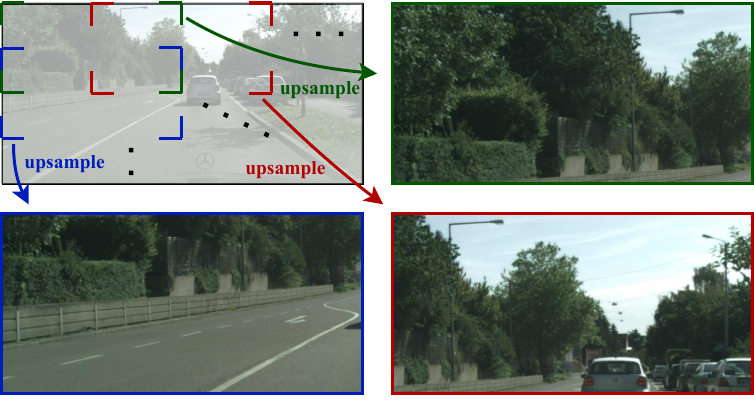}
    \vspace*{-.7cm}
    \caption{We perform multi-scale test-time augmentation with overlapping image crops to mitigate visual artifacts at the borders. Before feeding the crops to the task-specific networks, we upsample them to the original image size. In this figure, we illustrate the approach for scale $s=2$ and an image crop overlap of $z=2$.}
    \label{fig:multi-scale}
    \vspace*{-.6cm}
\end{figure}

{\parskip=3pt
\noindent\textit{Multi-Scale Prediction.}
During test-time, we perform both semantic segmentation and object boundary detection on multiple scales enabling our method to create more fine-grained predictions. In particular, we partition the input image of size $(w, h)$ into smaller areas of size
\begin{equation}
    w_s = w / s \, , \quad h_s = h / s \, ,
\end{equation}
where $s$ denotes the scale. Importantly, we propose to utilize overlapping image crops with strides
\begin{equation}
    r_{w, s} = \frac{w_s}{z} \, , \quad r_{h, s} = \frac{h_s}{z} \, .
\end{equation}
Unlike non-overlapping ensembles~\cite{kaeppeler2023spino}, our approach prevents sharp borders within the merged prediction that can result in visual artifacts. The parameter~$z$ defines the extent of the overlap, e.g., $z=2$ indicates that half of an image crop is overlapped by another crop.
We depict this method in \cref{fig:multi-scale} for scale $s=2$ and overlap $z=2$, yielding nine image crops. We upsample each crop to the input image size $(w, h)$ using bilinear interpolation and feed them through the respective head. Then, we downsample the generated feature maps to $(w_s, h_s)$ and place them in a combined feature map at the position corresponding to the input image crop. We repeat this procedure for each scale and average features of overlapping pixels. Finally, we merge the features from multiple scales using the mean value per pixel.
}


{\parskip=3pt
\noindent\textit{Panoptic Fusion and Refinement.}
We visualize the individual steps of our proposed panoptic fusion module in \cref{fig:overview}, starting with the previously described multi-scale prediction~(1).
First, we compute an affinity matrix $\mathbf{A}$ from the predicted soft boundary $\mathbf{\hat{B}}_\mathit{soft}$~(2) to be used for instance delineation~(9). We detail these steps in the next paragraph.
In step~(3), we obtain the binary boundary~$\mathbf{\hat{B}}$ after thresholding the class probabilities of each element $b^\mathit{soft}_{ij} \in \mathbf{\hat{B}}_\mathit{soft}$ with $\lambda_b$:
\begin{equation}
    b_{ij} = 
    \begin{cases}
        1 & \text{if } b^\mathit{soft}_{ij} > \lambda_b \\
        0 & \text{otherwise}
    \end{cases} \, , \ 
    b_{ij} \in \mathbf{\hat{B}}
\end{equation}
We further denoise $\mathbf{\hat{B}}$ by removing small boundaries.
Next, we extract boundaries between any two ``stuff'' and ``thing'' classes~(4) and add them to $\mathbf{\hat{B}}$. This enables us to find disconnected segments in the predicted semantic map~$\mathbf{\hat{S}}$. In detail, in step~(5) we perform connected component analysis (CCA) on $\mathbf{\hat{B}}$, followed by majority voting to update the pixel-wise predicted semantic classes~$\hat{y}_i$} of a segment~$\mathit{seg}$:
\begin{equation}
    \hat{y}_i = \argmax_{y \in \{1, \dots, n\}} \sum_{\hat{y}_j \in \mathit{seg}} \mathds{1} \left[ \hat{y}_j = y \right]
\end{equation}
In \cref{fig:overview}, this changes the burgundy colored pixels in the left vehicles to the \textit{car} class.
In step~(6), we again perform CCA but use the semantic predictions, i.e., we obtain a segment for each area in the image whose neighbors belong to a different semantic category. In the following, we separate these segments into ``stuff'' and ``thing'' segments. 
First, we iterate over the ``thing'' segments~(7) and set the semantic label to the \textit{ignore} class if the segment size is below a threshold, the boundary head does not predict a boundary for any of the segments' pixels, or the segment is fully surrounded by another thing class. While the second filter is inspired by ensemble learning, the third filter targets infeasible objects \textit{flying} in the scene. In \cref{fig:overview}, these filters remove the red pixels in the rear window of the center vehicle as well as the smaller \textit{car} segments.
Then, we iterate over the ``stuff'' segments~(8) and apply the same filters except for the boundary-based removal.
In step~(9), we fuse the semantic prediction with the detected object boundaries to obtain a panoptic map.
Finally, in step~(10), we propagate labels from the nearest neighbor to fill all previously created holes, i.e., pixels that were assigned the \textit{ignore} label. Note that although we only visualize \textit{car} and \textit{road} classes in \cref{fig:overview}, the final panoptic segmentation map~$\mathbf{\hat{P}}$ contains all valid categories.


{\parskip=3pt
\noindent\textit{Instance Separation.}
Here, we further detail steps~(2) and~(9) as numbered in \cref{fig:overview}. While CCA on the pixels assigned to the same ``thing'' class already yields disconnected image segments, the number and exact location of instances within a segment remains unknown. To delineate instances, we employ recursive two-way normalized cut (NCut)~\cite{shi2000ncut} to each image segment of a ``thing'' class.

In step (2), we first downsample the soft boundary map $\mathbf{\hat{B}}_\mathit{soft}$ to size $(w_b, h_b)$. Then, we compute a sparse affinity matrix $\mathbf{A} \in \mathbb{R}^{h_b \cdot w_b \times h_b \cdot w_b}$ based on the distance matrix $\mathbf{D}$with distances between pixels $p_i$ and $p_j$ defined as:
\begin{equation}
    d_{ij} = \max_{p_l \in \pline (p_i, p_j)} \mathbf{\hat{B}}_\mathit{soft}(p_l) \, ,
\end{equation}
where where the $\pline(\cdot, \cdot)$ operator is provided by the Bresenham algorithm. We convert the distances to affinities by taking the negative exponential:
\begin{equation}
    a_{ij} = e^{-\beta d_{ij}} \, ,
\end{equation}
where $a_{ij} \in \mathbf{A}$ and \mbox{$d_{ij} \in \mathbf{D}$}. The decay rate parameter $\beta$ controls the sensitivity of the affinity to changes in the distance. We interpret $\mathbf{A}$ as a weighted radius neighborhood graph with nodes and edges representing image pixels and affinities between neighboring pixels, respectively.

In step (9), we mask those elements in $\mathbf{A}$ that are not part of the current image segment and apply recursive NCut to cut the segment into instances. The objective of NCut is to minimize the cost of dividing a graph into two separate sub-graphs. The continuous solution of NCut is given by the eigenvector $v$ that corresponds to the second smallest eigenvalue $\lambda$ of the generalized eigenvalue problem:
\begin{equation}
    (\mathbf{C}-\mathbf{A}) x = \lambda \mathbf{C} x \, ,
\end{equation}
where $\mathbf{C}$ is a diagonal matrix with diagonal elements $c_{ii} = \sum_{j} \mathbf{A}_{ij}$. The solution $v$ is a continuous bipartition of the segment. If the cost of a cut, represented by $\lambda$, is less than a threshold and a stability criterion is fulfilled, we apply the cut. Otherwise, we stop the recursion. The stability criterion~\cite{shi2000ncut} measures the degree of smoothness in $v$. Formally, if the ratio between the minimum and the maximum value of a histogram of $v$ is less than a threshold, the criterion holds. Thus, we do not cut if there is high uncertainty in the solution $v$. To finally apply the cut, we search for the splitting point that minimizes the NCut cost to bipartition $v$ into two segments. We then recursively apply this procedure to both segments to find additional instances. If the size of a segment is below a threshold, we remove this segment and set its semantic label to the \textit{ignore} class. While using CCA~\cite{kaeppeler2023spino} fails for non-closed object boundaries, NCut is robust to small gaps and noise present in the boundary map.
}


\begin{figure}[t]
    \centering
    \includegraphics[width=\linewidth]{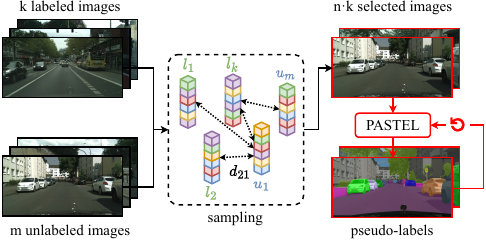}
    \vspace*{-.7cm}
    \caption{During self-training, we extract feature vectors $\{l_1, l_2, \dots, l_k\}$ of the labeled images as well as feature vectors $\{u_1, \dots, u_m\}$ of unlabeled images. Since the performance of \net is better on those unlabeled images that are more similar to the samples in the training set, we leverage the cosine similarity as distance measure $d_{ij}$ for image sampling.}
    \label{fig:self-training}
    \vspace*{-.6cm}
\end{figure}

\subsection{Iterative Self-Training}

Due to leveraging multi-scale predictions followed by refinements made by our panoptic fusion module, the final output of \net is of higher quality compared to the initial single-scale predictions, thus motivating iterative self-training. Because of the design of the fusion module, this enhancement mostly applies to the semantic output and is negligible for the boundary prediction. Therefore, we employ self-training only for the semantic segmentation head.

In particular, we propose to select unlabeled images in a feature-driven manner, as illustrated in \cref{fig:self-training}. First, we generate feature representations of the selected $k$ images with ground truth annotations as well as of a set of $m$ unlabeled images using the DINOv2~\cite{oquab2023dinov2} backbone. Then, we query the $n$ nearest neighbors from the unlabeled set for each image in our training set. Inspired by place recognition in visual SLAM~\cite{voedisch2023continual}, we utilize the cosine similarity between feature vectors as a similarity measure for images.
\begin{equation}
    \text{sim}_{\cos}(\mathbf{I_a}, \mathbf{I_b}) = \cos \left( \text{feat}(\mathbf{I_a}), \text{feat}(\mathbf{I_b}) \right) \, ,
\end{equation}
where $\mathbf{I}$ denotes an image with corresponding features $\text{feat}(\mathbf{I})$. We observe that the semantic predictions for these similar images are better than the predictions of a randomly sampled image and can hence bootstrap the semantic head. Please see \cref{ssec:ablations} for a quantitative argument.

Next, we use \net to create panoptic predictions for the sampled $n \cdot k$ images and treat them as pseudo-labels. We continue the training of the semantic segmentation head by constructing batches that contain both a ground truth annotation and a pseudo-labeled image. We use the same loss as in \cref{eqn:sem-loss} for the ground truth sample but set $t_K = 1.0$ for the pseudo-labeled image.
\looseness=-1

\section{Experimental Evaluation}
\label{sec:evaluation}

We demonstrate that our \net method outperforms previous label-efficient segmentation techniques while requiring significantly fewer annotations. We further showcase that using \net as a plugin can render state-of-the-art segmentation models label-efficient. In extensive ablation studies, we analyze the various design choices.


\subsection{Datasets and Implementation Details}
We present results on three diverse datasets:
First, the Cityscapes dataset~\cite{cordts2016cityscapes} provides RGB images and high-quality panoptic annotations with 19 classes for urban driving.
Second, the Pascal VOC 2012 dataset~\cite{everingham2010pascal} was originally proposed as an object detection benchmark and has been substantially extended by SBD~\cite{hariharan2011semantic}. Concerning panoptic segmentation, the dataset comprises 20 ``thing'' classes and a single ``stuff'' class representing the background.
Finally, the PhenoBench dataset~\cite{weyler2024phenobench} comprises several segmentation tasks for the agricultural domain. We apply our method to the leaf instance segmentation challenge. In stark contrast to autonomous driving, agricultural robotics lacks large-scale datasets underlining the importance of highly label-efficient approaches.
In our experiments, we select $k$ images from the \texttt{train} split of the respective dataset and report results on the \texttt{val} split. On PhenoBench, we provide further metrics for the \texttt{test} split.
If not noted otherwise, we train both heads for 150 epochs on an Nvidia RTX A6000 GPU taking \SI{11.5}{\minute} and \SI{16.7}{\minute} for semantic segmentation and object boundary estimation, respectively. The inference time with our default settings is approx. \SI{200}{\second} per image. We discuss real-time deployment in the last paragraph of \cref{ssec:results}.


\begin{table}[t]
\scriptsize
\centering
\caption{Image Segmentation on Cityscapes}
\vspace*{-0.2cm}
\label{tab:baselines-cityscapes}
\setlength\tabcolsep{5.0pt}
\begin{threeparttable}
    \begin{tabular}{ l@{\hskip 3.0pt}l | c | c | cc }
        \toprule
        \multicolumn{2}{l|}{\textbf{Method}} & \textbf{Backbone} & \textbf{Supervision} & mIoU & PQ \\
        \midrule
        1a) & Mask2Former~\cite{cheng2022mask2former} & Swin-L & $\mathcal{L}$ & 82.9 & 66.6 \\
        \greyrule
        2a) & PiCIE\textsuperscript{\textdagger}~\cite{hyun2021picie} & ResNet-18 & $\mathcal{U}$ & 13.8 & -- \\
        2b) & STEGO\textsuperscript{\textdagger}~\cite{hamilton2022stego} & DINO & $\mathcal{U}$ & 38.0 & -- \\  
        \greyrule
        3a) & ST++~\cite{yang2022stpp} & ResNet-50 & $\mathcal{L}_{100}$ + $\mathcal{U}$ & 61.4 & -- \\
        \greyrule
        4a) & ST++~\cite{yang2022stpp} & ResNet-50 & $\mathcal{L}_{100}$ & 55.1 & -- \\
        4b) & Hoyer~\textit{et~al.}~\cite{hoyer2021three} & ResNet-101 & $\mathcal{L}_{100}^*$ & 62.1 & -- \\
        4c) & Hoyer~\textit{et al.}\textsuperscript{\textdaggerdbl} & DINOv2 ViT-B & $\mathcal{L}_{10}$ & 46.4 & -- \\
        4d) & ST++\textsuperscript{\textdaggerdbl} & DINOv2 ViT-B & $\mathcal{L}_{10}$ & 53.0 & -- \\
        4e) & PanopticDeepLab\textsuperscript{\textdagger}~\cite{cheng2020panoptic} & DINOv2 ViT-B & $\mathcal{L}_{10}$ & 49.4 & 20.6 \\
        4f) & Mask2Former\textsuperscript{\textdaggerdbl} & Swin-L & $\mathcal{L}_{10}$ & 50.7 & 29.2 \\
        4g) & SPINO~\cite{kaeppeler2023spino} & DINOv2 ViT-B & $\mathcal{L}_{10}$ & 61.2 & 36.5 \\
        \greyrule
        5a) & \net (\textit{ours}) & DINOv2 ViT-L & $\mathcal{L}_{100}$ & 75.5 & 50.7 \\
        5b) & \net (\textit{ours}) & DINOv2 ViT-S & $\mathcal{L}_{100}^*$ & 64.2 & 41.0 \\
        5c) & \net (\textit{ours}) & DINOv2 ViT-B & $\mathcal{L}_{10}$ & 63.3 & 41.3 \\
        5d) & \net (\textit{ours}) & DINOv2 ViT-B & $\mathcal{L}_{10} + \mathcal{U}_{50}$ & 64.8 & 42.4 \\
        \bottomrule
    \end{tabular}
    \footnotesize
     Supervision methods $\mathcal{L}$ and $\mathcal{U}$ denote labeled and unlabeled data. If a subscript $k$ is specified, only $k$ images were used for training.
    The metrics of $\mathcal{L}^*_{100}$ are averaged over the same three fixed sets~\cite{hoyer2021three}.
    \textdagger:~Values are taken from SPINO~\cite{kaeppeler2023spino}. \textdaggerdbl:~Baselines trained by us.
\end{threeparttable}
\vspace*{-0.4cm}
\end{table}

\begin{table}[t]
\scriptsize
\centering
\caption{Image Segmentation on Pascal VOC 2012}
\vspace*{-0.2cm}
\label{tab:baselines-pascal}
\setlength\tabcolsep{5.0pt}
\begin{threeparttable}  
    \begin{tabular}{ l@{\hskip 3.0pt}l | c | c | cc }
        \toprule
        \multicolumn{2}{l|}{\textbf{Method}} & \textbf{Backbone} & \textbf{Supervision} & mIoU & PQ \\
        \midrule
        1a) & Panoptic FCN~\cite{li2023panopticfcn} & ResNet-50 & $\mathcal{L}$ & 80.2 & 67.9 \\
        \greyrule
        2a) & MaskContrast~\cite{gansbeke2021maskcontrast} & ResNet-50 & $\mathcal{U}$ & 35.0 & -- \\
        2b) & MaskDistill~\cite{gansbeke2022maskdistill} & ResNet-50 & $\mathcal{U}$ & 48.9 & -- \\
        \greyrule
        3a) & U\textsuperscript{2}PL~\cite{wang2022u2pl} & ResNet-101 & $\mathcal{L}_{92}$ + $\mathcal{U}$ & 68.0 & -- \\
        3c) & ST++~\cite{yang2022stpp} & ResNet-50 & $\mathcal{L}_{92}$ + $\mathcal{U}$ & 65.2 & -- \\
        \greyrule
        4a) & U\textsuperscript{2}PL~\cite{wang2022u2pl} & ResNet-101 & $\mathcal{L}_{92}$ & 45.8 & -- \\
        4b) & ST++~\cite{yang2022stpp} & ResNet-50 & $\mathcal{L}_{92}$ & 50.7 & -- \\
        4c) & ST++\textsuperscript{\textdaggerdbl} & DINOv2 ViT-B & $\mathcal{L}_{20}$ & 52.8 & -- \\
        \greyrule
        5a) & \net (\textit{ours}) & DINOv2 ViT-L & $\mathcal{L}_{92}$ & 71.1 & 47.3 \\
        5b) & \net (\textit{ours}) & DINOv2 ViT-B & $\mathcal{L}_{20}$ & 60.6 & 37.0 \\
        5c) & \net (\textit{ours}) & DINOv2 ViT-B & $\mathcal{L}_{20} + \mathcal{U}_{100}$ & 62.5 & 39.5 \\
        \bottomrule
    \end{tabular}
    \footnotesize
    Supervision methods $\mathcal{L}$ and $\mathcal{U}$ denote labeled and unlabeled data, respectively. If a subscript $k$ is specified, only $k$ images were used for training. \textdaggerdbl:~Baseline trained by us.
\end{threeparttable}
\vspace*{-.3cm}
\end{table}

\begin{table}[t]
\scriptsize
\centering
\caption{PhenoBench Leaf Instance Segmentation}
\vspace*{-0.2cm}
\label{tab:baselines-phenobench}
\setlength\tabcolsep{3.25pt}
\begin{threeparttable}
    \begin{tabular}{ l@{\hskip 3.0pt}l | c | c | c c }
        \toprule
        \multicolumn{2}{l|}{\textbf{Method}} & \textbf{Backbone} & \textbf{Supervision} & PQ (\texttt{val}) & PQ (\texttt{test}) \\
        \midrule
        1a) & Mask R-CNN \cite{kaiming2017maskrcnn} & ResNet-50 & $\mathcal{L}$ & 61.5 & 59.7 \\
        \greyrule
        2a) & Mask R-CNN\textsuperscript{\textdaggerdbl} & ResNet-50 & $\mathcal{L}_{15}$ & 41.5 & 38.2 \\
        2b) & PASTEL (\textit{ours}) & DINOv2 ViT-B & $\mathcal{L}_{15}$ & 51.7 & 49.0 \\ 
        \bottomrule
    \end{tabular}
    \footnotesize
    In 2a) and 2b), we used 15 images for training.
    \textdaggerdbl:~Baseline trained by us.
\end{threeparttable}
\vspace*{-.5cm}
\end{table}


\subsection{Panoptic Segmentation Results}
\label{ssec:results}

For evaluation of \net and other baseline methods, we report mIoU and PQ metrics~\cite{kirillov2019panoptic}. While the mIoU only refers to semantic segmentation, the PQ measures the quality of performing panoptic segmentation.


{\parskip=3pt
\noindent\textit{Comparison With Related Works.}
We evaluate \net with respect to previous label-efficient methods for semantic and panoptic segmentation. Additionally, we list metrics of fully supervised state-of-the-art approaches for panoptic segmentation on the respective dataset. Importantly, most prior techniques consider a minimum of 100 labeled images on Cityscapes and 92 on Pascal VOC, which is ten and five times more than our intended use case. Therefore, we report results for three different scenarios: First, with a minimum number of annotated images showcasing the label efficiency of our method. Second, we extend the setting to semi-supervision by including a few unlabeled images via our proposed self-training scheme. Finally, based on the encouraging results in \cref{ssec:ablations}, we investigate the potential performance of \net when increasing the available compute resources and the number of annotated images, resembling the setup from previous label-efficient methods.

\begin{figure*}[t]
    \centering
    \includegraphics[width=\linewidth]{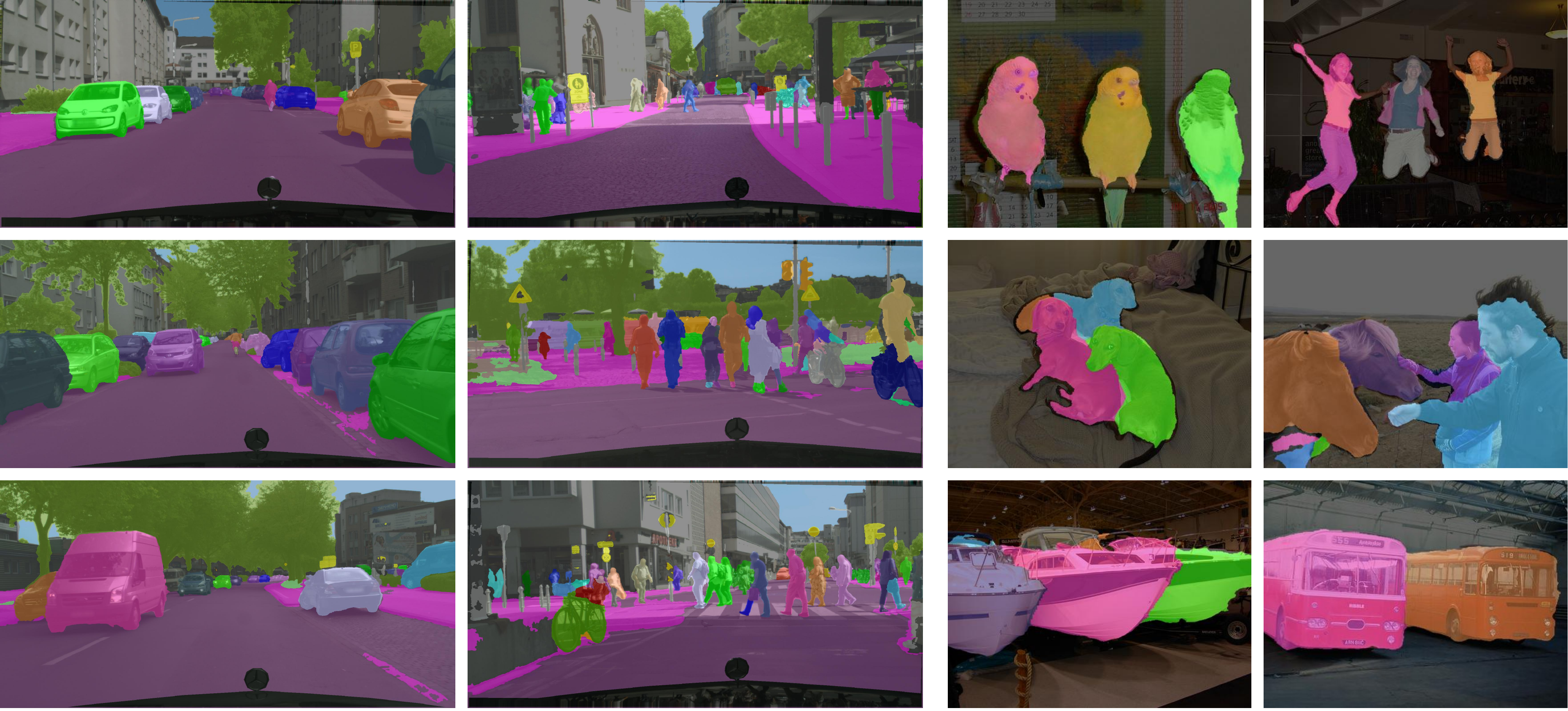}
    \vspace*{-.7cm}
    \caption{We provide qualitative results on both Cityscapes (\textit{left}) and Pascal VOC (\textit{right}) for examples taken from the respective \texttt{val} split. The depicted results are generated by \net based on the semi-supervised setup, i.e., $\mathcal{L}_k + \mathcal{U}_{n \cdot k}$.}
    \label{fig:qualitative-results}
    \vspace*{-.4cm}
\end{figure*}

\begin{figure}[t]
    \centering
    \includegraphics[width=\linewidth]{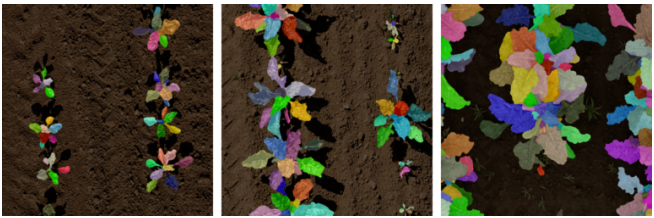}
    \vspace*{-.7cm}
    \caption{Qualitative results for the PhenoBench leaf instance segmentation challenge including different growth stages of the crops.}
    \label{fig:qualitative-results-phenobench}
    \vspace*{-.5cm}
\end{figure}

In \cref{tab:baselines-cityscapes}, we report results for the Cityscapes dataset. We first address the main target use case by allowing only ten annotated images for training. Our \net~(5c) achieves remarkable \SI{63.3}{\percent}~mIoU and \SI{41.3}{\percent}~PQ corresponding to an increase of $+2.1$~mIoU and $+4.8$~PQ to the recent SPINO~\cite{kaeppeler2023spino} (4g). Notably, the state-of-the-art works Mask2Former~\cite{cheng2022mask2former} (4f) and PanopticDeepLab~\cite{cheng2020panoptic} adapted with a DINOv2~\cite{oquab2023dinov2} backbone (4e) perform significantly worse revealing the need for specifically designed methods for extreme label efficiency. We further evaluate the work from Hoyer~\textit{et~al.}~\cite{hoyer2021three} (4c) and ST++~\cite{yang2022stpp} (4d) but replace the backbones with DINOv2 ViT-B to eliminate their impact. Besides the lack of instance predictions, this results in $-16.9$ and $-10.3$~mIoU compared to \net.
After employing our proposed self-training strategy (5d), the improvement over the concurrent SPINO~\cite{kaeppeler2023spino} further increases to $+3.6$~mIoU and $+5.9$~PQ, also yielding higher metrics than the semi-supervised ST++~\cite{yang2022stpp} (3a) that is trained with more labels.
For the third case, we use \net with a DINOv2 ViT-S backbone (86M param.) (5b) and compare it with the ResNet-101-based~\cite{he2016resnet} Hoyer~\textit{et~al.}~\cite{hoyer2021three} (45M param.) (4b) when training on the same 100 annotations. Our approach yields $+2.1$~mIoU plus instance predictions.
Finally, we show the potential of \net with a DINOv2 ViT-L backbone (5a) that expands the increase to $+13.4$~mIoU. Notably, this reduces the gap to the fully supervised Mask2Fomer~\cite{cheng2022mask2former} (1a) to $7.4$~mIoU and $15.9$~PQ while using \SI{3.4}{\percent} of the labels.

In \cref{tab:baselines-pascal}, we repeat similar experiments on the Pascal VOC dataset. When using only 20 annotated images, \net~(5b) yields \SI{60.6}{\percent}~mIoU and \SI{37.0}{\percent}~PQ, outperforming previous densely supervised methods with limited samples. Compared to ST++~\cite{yang2022stpp} with a DINOv2 ViT-B backbone (4c), the metrics represent $+7.8$~mIoU when trained with the same images. Similar to Cityscapes, self-training further increases the performance of \net~(5c). For the third setup, \net~(5a) achieves an increase of $+20.4$~mIoU versus the supervision-only baseline reported in ST++~\cite{yang2022stpp} (4b) when using the same number of annotations. On Pascal VOC, we can reduce the gap to the fully supervised Panoptic~FCN~\cite{li2023panopticfcn} (1a) to $9.1$~mIoU and $20.6$~PQ while using only \SI{0.8}{\percent} of the labels.

In \cref{tab:baselines-phenobench}, we report results for the PhenoBench leaf instance segmentation task. Separating leaves is essential to assess the growth stage of crops and to detect diseases. However, annotating a conventional training set with hundreds of images is infeasible. Thus, we demonstrate that \net~(2b) significantly outperforms the best-performing baseline Mask~R-CNN~\cite{kaiming2017maskrcnn} from the dataset's benchmark, when only \SI{1.1}{\percent} of the labeled images are available (2a).

Finally, we provide qualitative results in \cref{fig:qualitative-results} and \cref{fig:qualitative-results-phenobench}.
Albeit the complexity of the Cityscapes scenes, we observe that \net segments most \textit{car} instances correctly. Further, the more challenging \textit{pedestrians} are generally assigned the correct semantic class with minor over-segmentation of smaller body parts.
The images of Pascal VOC are usually less complex and contain a smaller variety of classes within a single image. In the depicted results, \net successfully separates instances of the same semantic class even in difficult scenes.
For the PhenoBench leaf instance segmentation challenge, the predictions of \net remain stable over the different growth stages of the crops.
}


\begin{table}
\scriptsize
\centering
\caption{Evaluation of Mask2Former with ResNet-50}
\vspace*{-0.2cm}
\label{tab:pseudo-labels}
\begin{threeparttable}  
    \begin{tabular}{ c c | cccc }
        \toprule
        \textbf{Training data} & \textbf{Split} & mIoU & PQ & SQ & RQ  \\
        \midrule
        Ground truth & \texttt{train} & 75.2 & 59.2 & 81.0 & 71.9 \\
        Pseudo-labels & \texttt{train} & 63.2 & 44.0 & 76.5 & 54.6 \\
        Pseudo-labels & \texttt{train\_extra} & 64.6 & 44.8 & 76.5 & 55.8 \\
        \bottomrule
    \end{tabular}
    \footnotesize
\end{threeparttable}
\vspace*{-.6cm}
\end{table}

{\parskip=3pt
\noindent\textit{Usage as Pseudo-Label Generator.}
In this experiment, we leverage the panoptic predictions of \net as pseudo-labels to train a densely supervised panoptic segmentation model. In detail, we train Mask2Former~\cite{cheng2022mask2former} with a \mbox{ResNet-50}~\cite{he2016resnet} backbone using the official code for three different settings. First, we use the ground truth annotations of the \texttt{train} split of Cityscapes. Second, we generate panoptic pseudo-labels for the same data using \net with the $\mathcal{L}_{10} + \mathcal{U}_{50}$ setting. Finally, we add pseudo-labels on the \texttt{train\_extra} split showing how to leverage large unlabeled datasets with our method. On all pseudo-labels, we mask the static hood of the ego vehicle following previous works~\cite{chen2020naivestudent, kaeppeler2023spino}.
We report the performance on \texttt{val} data in \cref{tab:pseudo-labels}. Note that the numbers from the authors are slightly greater than our reproduced results, $+2.3$~mIoU and $+2.9$~PQ~\cite{cheng2022mask2former}, but do not include SQ and RQ metrics. Importantly, the panoptic metrics with \texttt{train} pseudo-labels exceed the results obtained directly with \net, i.e., training Mask2Former further bootstraps the panoptic segmentation performance without increasing the utilized number of human annotations.
When adding the \texttt{train\_extra} pseudo-labels, the panoptic segmentation scores can be further improved, achieving $+2.4$~PQ compared to the results of \net. In summary, this experiment not only demonstrates that \net can serve as a plugin rendering existing densely supervised segmentation models label-efficient but also enables real-time inference~\cite{cheng2022mask2former}.
}


\subsection{Ablations and Analysis}
\label{ssec:ablations}

We conduct extensive ablation studies on Cityscapes~\cite{cordts2016cityscapes} to analyze the effect of various components and hyperparameters. Throughout the tables, we highlight the parameters used in \cref{ssec:results} in gray. Except for the components analysis and the study on iterative self-training, we omit self-training to isolate the effect of a parameter. For further studies, e.g., image size, please refer to the supplementary material.


\begin{table}[t]
\scriptsize
\centering
\caption{Components Analysis}
\vspace*{-0.2cm}
\label{tab:ablation-components}
\begin{threeparttable}
    \begin{tabular}{ l | cccc }
        \toprule
        \textbf{Method} & mIoU & PQ & SQ & RQ  \\
        \midrule
        Scale 1 w. CCA inst. segm. & 57.3 & 30.0 & 70.7 & 38.8 \\
        \hspace{.5pt} + Multi-scale augmentation & 62.4 & 36.8 & 73.9 & 46.8 \\
        \hspace{.5pt} + Normalized cut & 62.7 & 38.5 & 73.9 & 49.0 \\
        \hspace{.5pt} + Refinement steps & 63.3 & 41.3 & 74.5 & 52.1 \\
        \rowcolor{Gray}
        \hspace{.5pt} + Self-training (1 iteration) & \textbf{64.8} & \textbf{42.4} & \textbf{75.7} & \textbf{53.1} \\
        \bottomrule
    \end{tabular}
    \footnotesize
    Each row also includes the components of all rows above.
\end{threeparttable}
\vspace*{-.2cm}
\end{table}

\begin{table}[t]
\scriptsize
\centering
\caption{Number and Selection of Labels}
\vspace*{-0.2cm}
\label{tab:ablation-number-labels}
\begin{threeparttable}
    \begin{tabular}{ c | cccc }
        \toprule
        \textbf{Count} & mIoU & PQ & SQ & RQ  \\
        \midrule
        5 & 57.2 & 36.6 & 69.4 & 46.3 \\
        \rowcolor{Gray}
        10 & 63.3 & 41.3 & 74.5 & 52.1 \\
        25 & 67.1 & 43.9 & 75.6 & 55.1 \\
        50 & 69.2 & 46.0 & 76.3 & 57.6 \\
        100 & \textbf{70.7} & \textbf{47.2} & \textbf{76.7} & \textbf{59.0} \\
        \greyrule
        10 (\textit{study}) & 64.0$\pm$3.3 & 40.8$\pm$1.4 & 74.3$\pm$1.7 & 51.6$\pm$2.1 \\
        \bottomrule
    \end{tabular}
    \footnotesize
\end{threeparttable}
\vspace*{-.4cm}
\end{table}

\begin{table}[t]
\scriptsize
\centering
\caption{Variants of the Backbone}
\vspace*{-0.2cm}
\label{tab:ablation-backbone}
\begin{threeparttable}
    \begin{tabular}{ c | cccc }
        \toprule
        \textbf{DINOv2} & mIoU & PQ & SQ & RQ  \\
        \midrule
        ViT-S/14 & 53.2 & 34.1 & 73.9 & 42.6 \\
        \rowcolor{Gray}
        ViT-B/14 & 63.3 & 41.3 & 74.5 & 52.1 \\
        ViT-L/14 & \textbf{66.2} & 44.2 & 75.0 & 55.9 \\
        ViT-g/14 & 65.8 & \textbf{44.7} & \textbf{75.2} & \textbf{56.2} \\
        \bottomrule
    \end{tabular}
    \footnotesize
\end{threeparttable}
\vspace*{-.2cm}
\end{table}

\begin{table}[t]
\scriptsize
\centering
\caption{Number of Self-Training Iterations}
\vspace*{-0.2cm}
\label{tab:ablation-number-iterations-self}
\begin{threeparttable}
    \begin{tabular}{ c | cccc }
        \toprule
        \textbf{Iterations} & mIoU & PQ & SQ & RQ  \\
        \midrule
        0 & 63.3 & 41.3 & 74.5 & 52.1 \\
        \rowcolor{Gray}
        1 & \textbf{64.8} & \textbf{42.4} & \textbf{75.7} & \textbf{53.1} \\
        2 & 62.9 & 41.4 & 75.2 & 51.8 \\
        \bottomrule
    \end{tabular}
    \footnotesize
    We used 50 images and 50 epochs.
\end{threeparttable}
\vspace*{-.4cm}
\end{table}

\begin{table}[t]
\scriptsize
\centering
\caption{Number of Self-Training Images}
\vspace*{-0.2cm}
\label{tab:ablation-number-images-self}
\begin{threeparttable}
    \begin{tabular}{ c | cccc }
        \toprule
        \textbf{Count} & mIoU & PQ & SQ & RQ  \\
        \midrule
        0 & 63.7 & 41.4 & 74.9 & 52.1 \\
        10 & 63.5 & 41.5 & 75.1 & 52.1 \\
        \rowcolor{Gray}
        50 & \textbf{64.8} & 42.4 & \textbf{75.7} & 53.1 \\
        100 & 63.7 & \textbf{42.6} & 75.3 & \textbf{53.4} \\
        200 & \textbf{64.8} & 42.5 & 75.3 & \textbf{53.4} \\
        \greyrule
        50 (\textit{random}) & 61.8$\pm$1.3 & 38.5$\pm$0.5 & 73.6$\pm$0.3 & 48.5$\pm$0.7 \\
        \bottomrule
    \end{tabular}
    \footnotesize
    We used one iteration of self-training with 50 epochs. For the randomly sampled images, we provide mean and standard deviation over three experiments.
\end{threeparttable}
\vspace*{-.55cm}
\end{table}


{\parskip=3pt
\noindent\textit{Components Analysis.}
We report the impact of the components of \net in \cref{tab:ablation-components}. The largest effect can be observed for multi-scale test-time augmentation, enabling our method to produce more detailed predictions. Next, we substitute instance delineation via CCA~\cite{kaeppeler2023spino} with recursive two-way normalized cut, improving the panoptic metrics. Employing the post-processing of our proposed panoptic fusion module increases both semantic and panoptic performance. Finally, we demonstrate that one iteration of self-training further boosts the performance by $+1.5$~mIoU and $+1.1$~PQ. In \cref{tab:ablation-number-images-self}, we further show that our proposed feature-driven similarity sampling performs significantly better than self-training with 50 randomly sampled images.
}


{\parskip=3pt
\noindent\textit{Choice of Labeled Images.}
To measure the effect of the selected images, we conduct a user study with four participants tasked to select ten RGB images covering all semantic classes while maximizing diversity. We report the mean and standard deviation in the bottom row of \cref{tab:ablation-number-labels} denoted by \textit{study}. The study shows that the performance of \net remains stable for different selections of training data.
Next, we evaluate the potential performance of \net if one would further increase the number of labeled training images, although this does not reflect the main goal of our work. Please note that we use the same images as SPINO~\cite{kaeppeler2023spino}, allowing for a direct comparison. In detail, \net achieves $+4.5$, $+5.4$, $+4.3$, $+5.1$, and $+4.3$~PQ compared to SPINO for an increasing label count from $L_5$ to $L_{100}$.
}


{\parskip=3pt
\noindent\textit{Backbone.}
We present results for different variants of \mbox{DINOv2}~\cite{oquab2023dinov2} in \cref{tab:ablation-backbone}. Note that we selected DINOv2 ViT-B/14 for our method as it compromises performance and computational feasibility. We observe that the larger backbone DINOv2 ViT-L/14 shows significant performance improvements, whereas increases due to DINOv2 ViT-g/14 are marginal. Similar to a study on image classification~\cite{oquab2023dinov2}, we hypothesize that the number of parameters of the \mbox{ViT-L/14} variant suffices to model the training data.
}


{\parskip=3pt
\noindent\textit{Iterative Self-Training.}
Finally, we conduct studies on the number of self-training iterations (\cref{tab:ablation-number-iterations-self}) and images sampled by our sampling strategy (\cref{tab:ablation-number-images-self}). We show that performing self-training once is sufficient to increase performance~\cite{wang2022freesolo}, while further iterations decrease the quality, most likely due to overfitting. For the image count, we sample the \mbox{$n = 5$} nearest neighbors for each annotated image in the training set and continue training the semantic head for 50 epochs. In \cref{tab:ablation-number-images-self}, we further report results for \mbox{$n \in \{0, 1, 10, 20 \}$}, where \mbox{$n=0$} corresponds to resuming the training without pseudo-labeled images. Note that our proposed similarity sampling yields significantly better results than self-training on randomly sampled images, shown in the bottom row.
}

\section{Conclusion}

In this work, we demonstrated that recent visual foundation models offer a powerful pretraining strategy for solving computer vision tasks in a label-efficient manner. In particular, we presented \net for label-efficient panoptic segmentation. Our method combines descriptive image features from a DINOv2~\cite{oquab2023dinov2} backbone with two lightweight heads for semantic segmentation and object boundary detection. It can be trained with as few as ten annotated images. We showed that our novel panoptic fusion module yields substantial performance improvements compared to previous works and illustrated how to further enhance the results using self-training with similar images.
Most notably, we demonstrated that \net sets the new state of the art for label-efficient segmentation by improving mIoU scores by $+13.4$ and $+20.4$ on Cityscapes and Pascal VOC datasets, respectively.
In future research, we aim to further close the gap to fully supervised methods paving the way for widespread application of panoptic segmentation without requiring large-scale annotated datasets.


\footnotesize
\bibliographystyle{IEEEtran}
\bibliography{references.bib}


\clearpage
\renewcommand{\baselinestretch}{1}
\setlength{\belowcaptionskip}{0pt}

\sisetup{output-exponent-marker=\ensuremath{\mathrm{e}}}

\begin{strip}
\begin{center}
\vspace{-5ex}

\textbf{\LARGE \bf
A Good Foundation is Worth Many Labels: \\\vspace{.5ex} Label-Efficient Panoptic Segmentation} \\
\vspace{3ex}

\Large{\bf- Supplementary Material -}\\
 \vspace{0.4cm}
 \normalsize{
Niclas Vödisch$^{1*}$,
Kürsat Petek$^{1*}$,
Markus Käppeler$^{1*}$,
Abhinav Valada$^{1}$, and
Wolfram Burgard$^{2}$}
\end{center}
\end{strip}

\setcounter{section}{0}
\setcounter{equation}{0}
\setcounter{figure}{0}
\setcounter{table}{0}
\setcounter{page}{1}
\makeatletter


\renewcommand{\thesection}{S-\Roman{section}}
\renewcommand{\thesubsection}{S-\arabic{subsection}}
\renewcommand{\thetable}{S-\Roman{table}}
\renewcommand{\thefigure}{S-\arabic{figure}}
\renewcommand{\theequation}{S-\arabic{equation}}

\let\thefootnote\relax\footnote{$^{*}$ Equal contribution.\\
 $^{1}$ Department of Computer Science, University of Freiburg, Germany.\\
 $^{2}$ Department of Eng., University of Technology Nuremberg, Germany.
}%
\normalsize


In this supplementary material, we provide additional experiments, ablation studies, and qualitative results. We conclude by discussing some limitations of our proposed \net approach.


\section{Pseudo-Labels for Efficient Pretraining}
\label{app:rationale}

In this section, we extend the evaluation of leveraging \net as a pseudo-label generator to enable label-efficient training of any existing panoptic segmentation model. As we demonstrate in \cref{ssec:results}, this process upgrades existing models from the classical dense supervision style, which requires many annotated images, to label-efficient training.
In \cref{app-tab:pseudo-labels}, we provide results from employing the predicted panoptic maps from both the \texttt{train} and \texttt{train\_extra} splits of Cityscapes~\cite{cordts2016cityscapes} as pseudo-labels to train Mask2Former~\cite{cheng2022mask2former} with a ResNet-50~\cite{he2016resnet} backbone. In contrast to \cref{tab:pseudo-labels}, we interpret this step only as pretraining and resume the training with the ground truth annotations from the \texttt{train} set.
In comparison to the results reported by the authors, who train only on the ground truth annotations, our label-efficient pretraining results in an increase of $+1.9$~mIoU and $+1.4$~PQ scores.

\begin{table}[h]
\scriptsize
\centering
\caption{Evaluation of Mask2Former with ResNet-50}
\vspace{-0.2cm}
\label{app-tab:pseudo-labels}
\begin{threeparttable}
    \begin{tabular}{ c c | cccc }
        \toprule
        \textbf{Label type} & \textbf{Data split} & mIoU & PQ & SQ & RQ  \\
        \midrule
        Ground truth\textsuperscript{\textdagger{}} & \texttt{train} & 77.5 & 62.1 & -- & -- \\
        \greyrule
        Pseudo-labels & \texttt{train\_extra} & 64.6 & 44.8 & 76.5 & 55.8 \\
        \hspace{.5pt} + Ground truth & \texttt{train} & 79.4 & 63.5 & 82.2 & 76.4 \\
        & & (+1.9) & (+1.4) \\
        \bottomrule
    \end{tabular}
    \footnotesize
    Supervised training results of Mask2Former~\cite{cheng2022mask2former} with a \mbox{ResNet-50}~\cite{he2016resnet} backbone. We pretrain the network on pseudo-labels generated by \net and then continue training on ground truth annotations. The results denoted by~\textdagger{} are reported by the authors~\cite{cheng2022mask2former}.
\end{threeparttable}
\vspace{-0.2cm}
\end{table}


\section{Ablations and Analysis}

In this supplementary section, we further extend the ablation studies provided in \cref{ssec:ablations}. We continue to highlight the parameters used in \cref{ssec:results} in gray. We further omit self-training to isolate the effect of the analyzed parameter.


{\parskip=3pt
\noindent\textit{Image Size.}
In this study, we ablate the effect of the image size on the overall performance. We report results in \cref{tab:ablation-image-size} for the full resolution as well as on scales~$\sfrac{1}{2}$ and~$\sfrac{1}{4}$. While scale~$\sfrac{1}{2}$ achieves decent metrics, the performance significantly decreases for scale~$\sfrac{1}{4}$.
}


{\parskip=3pt
\noindent\textit{Number of Epochs.}
In \cref{tab:ablation-number-epochs}, we evaluate the performance of \net after different numbers of training epochs showing a general trend of improvements until the metrics converge. Please note that we use a loss-based termination strategy during training, i.e., we explicitly do not select the number of epochs with the highest metrics in \cref{tab:ablation-number-epochs} as we consider them to represent test data.
}


\begin{table}[h]
\scriptsize
\centering
\caption{Image Size}
\vspace{-0.2cm}
\label{tab:ablation-image-size}
\begin{threeparttable}  
    \begin{tabular}{ c | cccc }
        \toprule
        \textbf{Image size} & mIoU & PQ & SQ & RQ  \\
        \midrule
        $\phantom{0}252\times\phantom{0}504$ & 56.7 & 31.9 & 71.4 & 41.0 \\
        $\phantom{0}504\times1008$ & \textbf{63.5} & 39.9 & 73.2 & 50.8 \\
        \rowcolor{Gray}
        $1022\times2044$ & 63.3 & \textbf{41.3} & \textbf{74.5} & \textbf{52.1} \\
        \bottomrule
    \end{tabular}
    \footnotesize
\end{threeparttable}
\end{table}

\begin{table}[h]
\scriptsize
\centering
\caption{Number of Epochs}
\vspace{-0.2cm}
\label{tab:ablation-number-epochs}
\begin{threeparttable}
    \begin{tabular}{ c | cccc }
        \toprule
        \textbf{Epoch} & mIoU & PQ & SQ & RQ  \\
        \midrule
         50 & 60.5 & 39.5 & 74.6 & 49.7 \\
        100 & 63.4 & 41.0 & 74.8 & 51.8 \\
        \rowcolor{Gray}
        150 & 63.3 & 41.3 & 74.5 & \textbf{52.1} \\
        200 & \textbf{63.7} & \textbf{41.4} & \textbf{74.9} & \textbf{52.1} \\
        250 & 63.0 & 41.0 & 75.1 & 51.5 \\
        \bottomrule
    \end{tabular}
    \footnotesize
\end{threeparttable}
\end{table}


\begin{figure*}
    \centering
    \includegraphics[width=\linewidth]{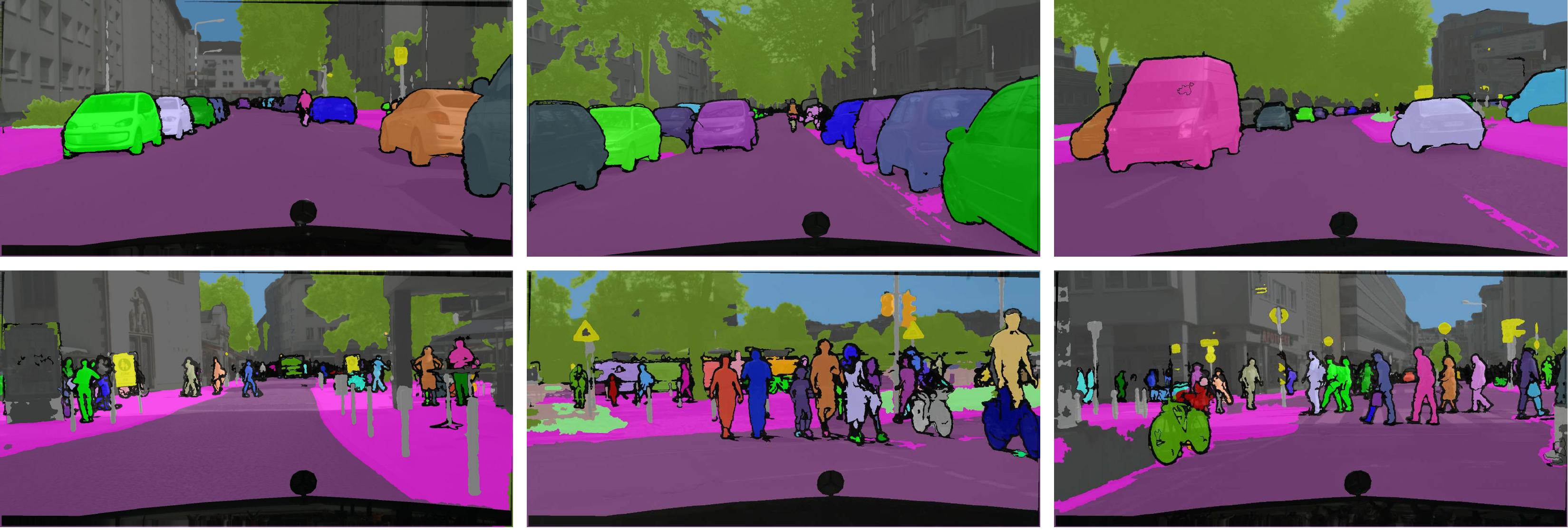}
    \vspace{-.5cm}
    \caption{Object boundaries predicted by \net based on the semi-supervised setup, i.e., $\mathcal{L}_{10} + \mathcal{U}_{50}$.}
    \label{app-fig:qualitative-results-cityscapes}
    \vspace{1cm}
\end{figure*}

\begin{figure*}
    \centering
    \includegraphics[width=\linewidth]{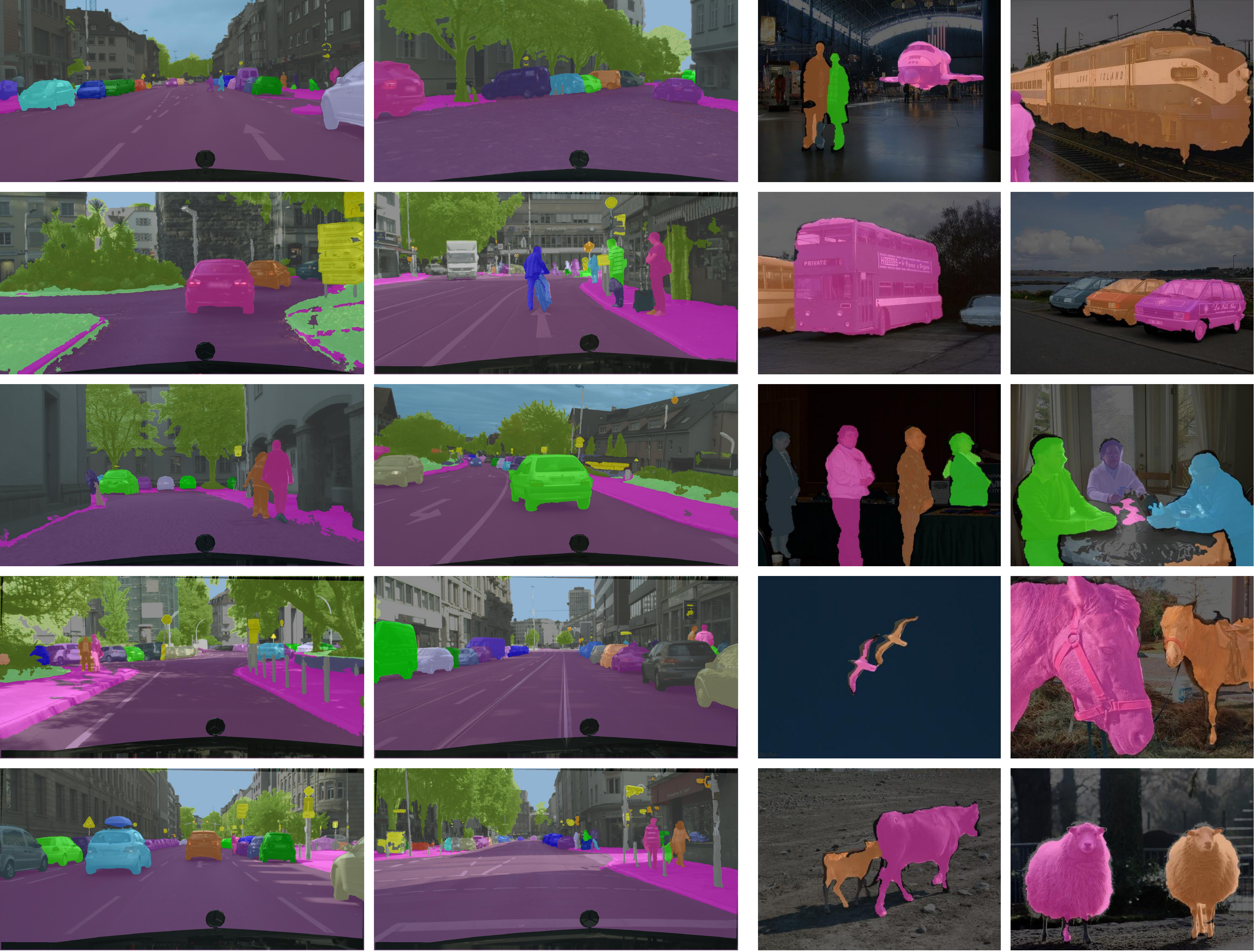}
    \vspace{-.5cm}
    \caption{Additional qualitative results on both Cityscapes (\textit{left}) and Pascal VOC (\textit{right}) datasets for examples taken from the respective \texttt{val} split. The depicted results are generated by \net based on the semi-supervised setup, i.e., $\mathcal{L}_k + \mathcal{U}_{n \cdot k}$.}
    \label{app-fig:qualitative-results}
\end{figure*}

\section{Discussion of Limitations}

\begin{figure}[t]
    \centering
    \includegraphics[width=.75\linewidth]{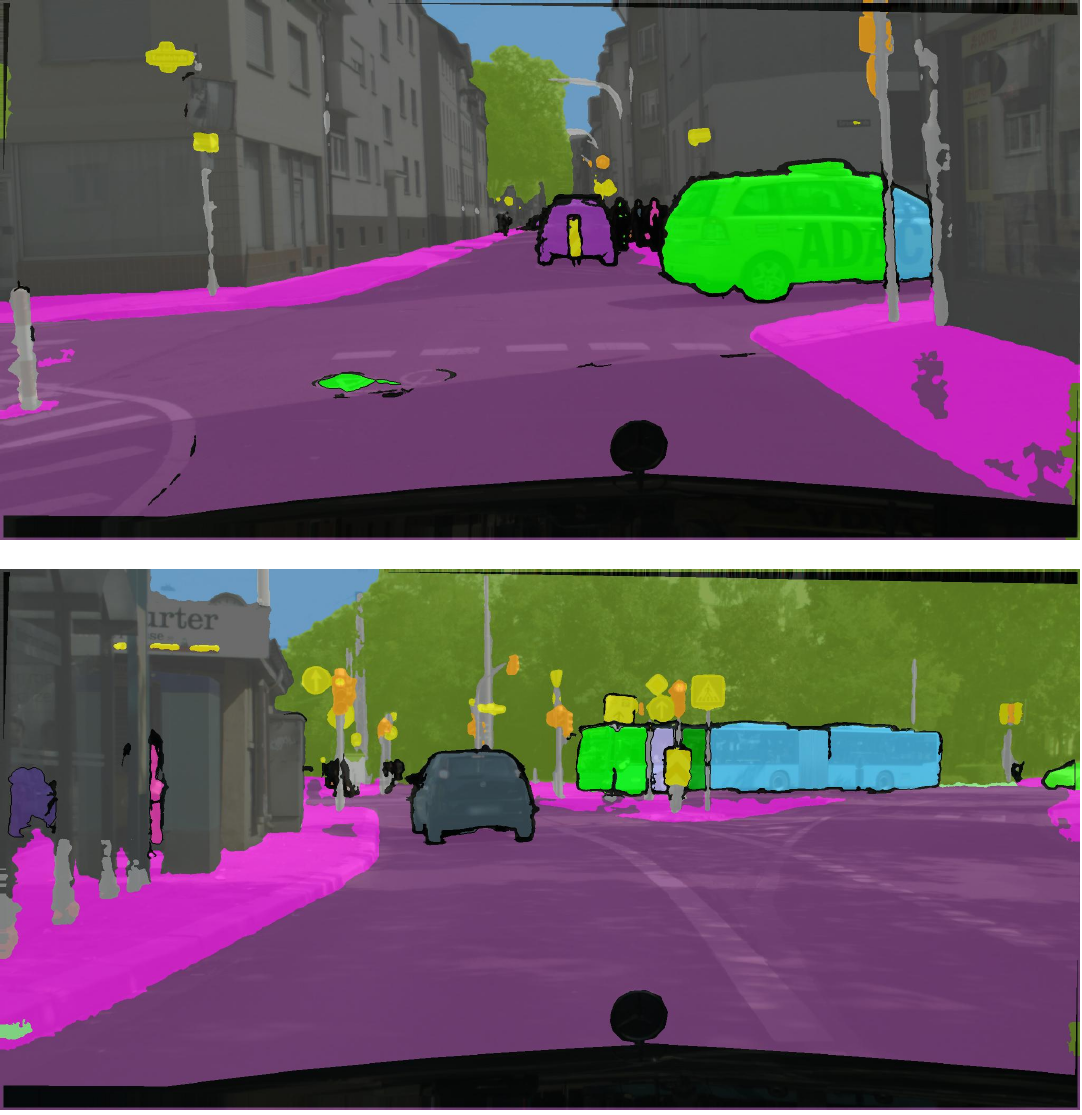}
    \vspace{-.1cm}
    \caption{A limitation of our method is that occluded objects are separated into multiple instances.}
    \label{app-fig:limitations-cityscapes}
\end{figure}

Our label-efficient segmentation approach is subject to two limitations.
First, since \net predicts the boundaries of objects based on the RGB input, areas that are separated in the 2D image space but belong to the same real-world objects cannot be assigned the same instance ID. We visualize examples of this failure case for occluded objects in \cref{app-fig:limitations-cityscapes}. In the upper image, the car on the right is cut into two instances due to occlusion by a traffic light. In the lower image, \net assigns four different instance IDs to different parts of the bus, which is occluded by multiple poles. Potential solutions to this limitation would be to employ amodal panoptic segmentation or a separate network head to predict the pixel offset similar to classical bottom-up methods. A key challenge of this future work will be to enable training with as few labeled images as utilized by \net.
Second, the minimum number of labeled images is constrained by the necessity for all considered classes to be present in the selected images. However, this constraint is not related to the proposed approach but is rather inherent in the need of the network to learn the notion of classes.

\section{Qualitative Results}
\label{app:qualitative-results}

\begin{figure}[b]
    \centering
    \includegraphics[width=\linewidth]{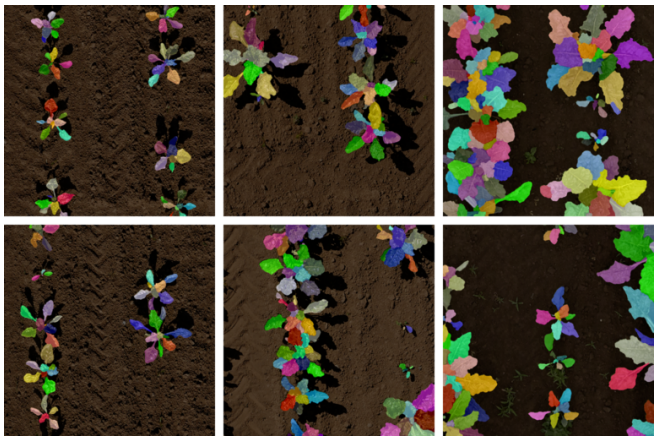}
    \vspace{-.6cm}
    \caption{Additional qualitative results for the PhenoBench leaf instance segmentation challenge including different growth stages of the crops.}
    \label{app-fig:qualitative-results-phenobench}
\end{figure}

In \cref{app-fig:qualitative-results}, we present additional qualitative results for both Cityscapes~\cite{cordts2016cityscapes} and Pascal VOC~\cite{everingham2010pascal} datasets. In \cref{app-fig:qualitative-results-phenobench}, we provide further qualitative results for the PhenoBench~\cite{weyler2024phenobench} leaf instance segmentation challenge.

{\parskip=3pt
\noindent\textit{Cityscapes.}
The examples from the Cityscapes dataset include scenes from all three cities within the \texttt{val} split, i.e., Frankfurt, Lindau, and Munster. Notably, our employed multi-scale prediction scheme allows for segmenting also more distanced details such as traffic signs.
In \cref{app-fig:qualitative-results-cityscapes}, we further visualize the predicted object boundary of the examples shown in \cref{fig:qualitative-results} of the main paper. As we observe in the pedestrian scenes, the over-segmentation of small body parts is caused by the predicted boundaries.
}

{\parskip=3pt
\noindent\textit{Pascal VOC.}
Since the majority of images in the Pascal VOC dataset contain only a single object, we deliberately show results on scenes with multiple objects including multiple instances of the same class as well as compositions of different ``thing'' classes. 
}

{\parskip=3pt
\noindent\textit{PhenoBench.}
Separating leaves is an important task for estimating the growth stage of plants and for detecting leaf diseases~\cite{weyler2024phenobench}. We provide results for all three stages contained in the \texttt{val} split of the dataset. Despite a drastic increase in scene complexity due to overlapping leaves, the prediction quality remains stable across the different stages.
}


\end{document}